# NQMIX: Non-monotonic Value Function Factorization for Deep Multi-Agent Reinforcement Learning

Quanlin Chen

## Abstract

Multi-agent value-based approaches recently make great progress, especially value decomposition methods. However, there are still a lot of limitations in value function factorization. In VDN, the joint action-value function is the sum of per-agent action-value function while the joint action-value function of QMIX is the monotonic mixing of per-agent action-value function. To some extent, QTRAN reduces the limitation of joint action-value functions that can be represented, but it has unsatisfied performance in complex tasks. In this paper, in order to extend the class of joint value functions that can be represented, we propose a novel actor-critic method called NQMIX. NQMIX introduces an off-policy policy gradient on QMIX and modify its network architecture, which can remove the monotonicity constraint of QMIX and implement a non-monotonic value function factorization for the joint action-value function. In addition, NQMIX takes the state-value as the learning target, which overcomes the problem in QMIX that the learning target is overestimated. Furthermore, NQMIX can be extended to continuous action space settings by introducing deterministic policy gradient on itself. Finally, we evaluate our actor-critic methods on SMAC domain, and show that it has a stronger performance than COMA and QMIX on complex maps with heterogeneous agent types. In addition, our ablation results show that our modification of mixer is effective.

## 1 Introduction

Multi-agent reinforcement learning (MARL) can be used to solve many real-world problems such as coordination of robot swarms [1] and autonomous cars [2]. But two crucial challenges remain open in MARL. One is credit assignment problem[3]: we only know the global reward returned from environment, but it is difficult for us to deduce the local reward for each agent. The other is that joint action space of MARL grows exponentially in the number of agents.

To cope with the credit assignment problem, VDN[4] and QMIX[5] learn to maximize the individual action-value function $Q_a$ conditioned on per-agent local observation history, and then modeling the joint action-value function $Q_{tot}$ as the mixing of individual value functions. In this way, global reward can be directly used to train the joint action-value function. To address the complexity of joint action space, QMIX's individual value function only takes each agent's local observation and action as input while its joint value function only takes per-agent action-value as input.

But QMIX still has some limitations: (1) The monotonicity constraint of QMIX between $Q_{tot}$ and each $Q_a$ will restricts the class of joint action-value functions that can be represented. (2) QMIX uses Q-learning to train the critic, which will overestimate the learning target [6]. (3) It's hard for QMIX to extend itself to continuous action space settings. Actually, if the action space is continuous, then $Q_a(\tau^a,\cdot)$ is continuous function. Besides, $Q_a(\tau^a,\cdot)$ is represented by neural networks, so $Q_a(\tau^a,\cdot)$ is a continuous nonconvex function that is hard for us to obtain the maximum point.

To address these limitations, we propose a novel actor-critic method called NQMIX: (1) NQMIX hybridizes off-policy policy-gradient and value-based methods so as to implement non-monotonic value function factorization for the joint action-value function. (2) NQMIX takes the state-value as the learning target to avoid overestimating the learning target. (3) we apply deterministic policy gradient on NQMIX to extend it to continuous action space settings.

## 2  Related Work

The simplest approach to extract policies in MARL is using Q-learning [7] to learn an individual action-value function independently for each agent (*independent Q-learning* [8]) or using actor-critic to learn an individual policy and critic for each agent (*independent actor-critic*[9]). However, this approach has the credit assignment problem and non-stationarity problem without representing interactions between agents.

Another approach is policy-based methods. They always learn a fully centralized value function which is used to train decentralized actor for each agent, which can represent interactions between agents and overcome the non-stationarity problem. For example, MADDPG[10] extends DDPG[11] to the MARL setting by introducing a centralized critic, but it still has the credit assignment problem. Similarly, COMA[9] also learns a centralized critic and then applies it to estimate a counterfactual baseline which can handle the credit assignment problem. Besides, COMA proposes the *counterfactual multi-agent* policy gradient, but it is an on-policy policy gradient which limits the speed of learning.

In addition, some approaches are value-based methods. They always model the centralized value function as mixing of individual value functions. VDN decomposes the centralized action-value function into the sum of per-agent action-value function while QMIX decomposes the centralized action-value function into the monotonic combination of per-agent action-value function. Moreover QTRAN[12] proposes a more general factorization than VDN or QMIX.

## 3  Background

A fully cooperative multi-agent task can be described as a Dec-POMDP[13] defined as a tuple $G = \langle A, S, U, P, Z, O, R \rangle$. $A = \{1, ..., n\}$ is a set of n agents. $S$ is true state space of the environment. $U = \times_{i \in A} U_i$ is the space of joint actions. $P$ is a transition function describing the probability $P(s'|s, \mathbf{u})$. $Z = \times_{i \in A} Z_i$ is the space of joint observations. $O$ is an observation function describing the probability $P(o_t|s_t, \mathbf{u}_{t-1})$. At each time step $t$, each agent $a \in A \equiv \{1, \dots, n\}$ choose an action $u_t^a$, which results in a joint action $\mathbf{u} \in U$. The joint action will cause a transition of environment described by transition function $P$. And the environment returns joint observations $\mathbf{o} = (o_t^1, o_t^2, \cdots, o_t^n)$ and a global reward $r \sim R(s, \mathbf{u})$. But each agent $a$ can only receive its local observation $o_t^a$. Besides, Dec-POMDP use $\tau^a$ to represent an action-observation history for agent $a$. At specific time step $t$, $\tau_t^a = (u_0^a, o_1^a \dots, u_{t-1}^a, o_t^a)$. And then a stochastic policy $\pi^a$ for each agent $a$ is modeled as a mapping from action-observation history $\tau^a$ to the probability of action $u^a$. In other words, $\pi^a(u^a|\tau^a, \boldsymbol{\theta}): T \times U \to [0,1]$.

**QMIX** is a value-based method that learns a monotonic joint action-value function $Q_{tot}$ as a mixing of per-agent action-value. First, QMIX estimates per-agent action-value by a *GRU network* that takes local action-observation history $\tau^a$ as input. Per-agent action-value function is denoted by $Q_a(\tau^a, \cdot)$. Each agent $a$ choose greedy actions with respect to its $Q_a(\tau^a, \cdot)$ and get the $Q_a(\tau^a, u^a)$. A *mixing network* with non-negative weights is responsible for combining $Q_a(\tau^a, u^a)$ into $Q_{tot}$ monotonically. And non-negative weights ensure $\frac{\partial Q_{tot}}{\partial Q_a} \geq 0, \forall a \in A$, which guarantees that

$$\underset{\mathbf{u}}{\operatorname{argmax}} Q_{tot}(\boldsymbol{\tau}, \mathbf{u}) = \begin{bmatrix} \underset{u^1}{\operatorname{argmax}} Q_1(\tau^1, u^1) \\ \vdots \\ \underset{u^n}{\operatorname{argmax}} Q_1(\tau^n, u^n) \end{bmatrix} \quad (1)$$

**Policy Gradient[14]** is a policy-based method that learns *parameterized policy* that can select actions without consulting a value function. We use $\boldsymbol{\theta} \in R^d$ to represent the policy's parameter vector. Then we write $\pi(u|s, \boldsymbol{\theta}) = Pr(U_t = u | S_t = s, \boldsymbol{\theta}_t = \boldsymbol{\theta})$ as *parameterized policy*. The *return* $G_t$ is the sum of discounted reward from time-step $t$, $G_t = \sum_{k=t}^{\infty} \gamma^{k-t} r(s_k, u_k)$ where $0 < \gamma < 1$. Value functions are defined to be the expected *return*, $V^\pi(s) = E(G_t|S_t = s; \pi)$ and $Q^\pi(s, u) = E(G_t|S_t = s, U_t = u; \pi)$. The agent's goal is to obtain a policy which maximizes the *return* from the start state, denoted by the performance objective $J(\pi) = V^\pi(s_0)$. From the *policy gradient theorem*[14], we know that

$$\nabla_\theta J(\pi) = \sum_s d^\pi(s) \sum_u \nabla_\theta \pi(u|s, \boldsymbol{\theta}) Q^\pi(s, u) \quad (2)$$

where $d^\pi(s) = \sum_{t=0}^{\infty} \gamma^t Pr\{S_t = s|s_0, \pi\}$.

**Off-policy actor-critic[15]** gives the *off-policy policy-gradient*

$$\nabla_\theta J(\boldsymbol{\theta}) \approx g(\boldsymbol{\theta}) = \sum_s d^b(s) \sum_u \nabla_\theta \pi(u|s, \boldsymbol{\theta}) Q^\pi(s, u) \quad (3)$$

where $b$ denotes the *behavior* policy that is used to generate behavior and $\pi$ denotes the *target* policy that is evaluated and improved.

# 4 Methodology

## 4.1 Policy Gradient for agent a

The monotonicity constraint is to ensure the consistency of greedy policy, so a natural thought is to replace greedy policy by parameterized policy. But there are more problems to solve with the introduction of actor, such as the credit assignment.

Conventional single-agent policy gradient can't solve the credit assignment, such as REINFORCE[16] whose form is as follows:

$$\theta_{t+1} = \theta_t + \alpha \gamma^t G_t \frac{\nabla_\theta \pi(u_t|s_t, \boldsymbol{\theta})}{\pi(u_t|s_t, \boldsymbol{\theta})} \quad (4)$$

where $G_t$ is difficult for us to compute, because we only know the global reward $r$, but we don't know the local reward for each agent. Although COMA can cope with the credit assignment problem in which the gradient is:

$$g = E_\pi[\nabla_\theta \log \pi^a(u^a|\tau^a) A^a(s, \mathbf{u})] \quad (5)$$

$$A^a(s, \mathbf{u}) = Q(s, \mathbf{u}) - \sum_{u'^a} \pi^a(u'^a|\tau^a) Q(s, (\mathbf{u}^{-a}, u'^a)) \quad (6)$$

yet COMA must be trained on-policy, which prevents us from using replay buffer and limits the learning speed.

In order to learn off-policy, we use the *all-actions* method from the exercise 13.2 of [16] (my solution is at Appendix A), in which we rewrite Equation 2 and 3 as an expectation when there's a discount $\gamma$:

$$\nabla_\theta J(\pi) = \sum_{t=0}^{\infty} \gamma^t E_s \left[ \sum_u \nabla_\theta \pi(u|s, \boldsymbol{\theta}) Q^\pi(s, u) \right] \quad (7)$$

Equation 7 not only overcomes the credit assignment problem as it doesn't refer to the global reward, but also is an off-policy policy gradient because we only instantiate expectation using the sampled $s_t \sim b$ rather than instantiate expectation using the sample action $\mathbf{u}$. And then we apply this formula to perform gradient-ascent algorithm. In detail, we instantiate the expectation using the sampled $\tau_t^a$ and get stochastic gradient-ascent formula:

$$\theta_{t+1} = \theta_t + \alpha \gamma^t \sum_u \nabla \pi(u|\tau_t^a) Q^\pi(\tau_t^a, u) \quad (8)$$

In addition, if the action space is continuous or very large, then we can use deterministic policy gradient[17]:

$$\nabla_\theta J(\mu_\theta) = E_{s \sim \rho^\mu} [\nabla_\theta \mu_\theta(s) \nabla_u Q^\mu(s, u)|_{u=\mu_\theta(s)}] \quad (9)$$

We also instantiate this expectation using sampled $\tau_t^a$ and get another gradient-ascent formula:

$$\theta_{t+1} = \theta_t + \alpha\gamma^t \nabla_\theta \mu_\theta(\tau_t^a) \nabla_u Q^\mu(s,u)|_{u=\mu_\theta(\tau_t^a)} \qquad (10)$$

## 4.2 Remove the monotonicity constraint

The policy of QMIX is a greedy policy, so it needs monotonicity to ensure that each agent always contributes to joint action-value. But when we learn a parameterized policy, it is possible for us to choose to increase or decrease per-agent performance based on the sign of $\frac{\partial Q_{tot}}{\partial Q_a}$. In detail, for per agent $a$, if $\frac{\partial Q_{tot}}{\partial Q_a} > 0$, then we should increase the performance $J(\pi^a)$ of the agent $a$, which means that we should perform gradient-ascent algorithm:

$$\theta_{t+1} = \theta_t + \alpha\gamma^t \sum_u \nabla \pi(u|\tau_t^a) Q^\pi(\tau_t^a, u) \qquad (11)$$

On the contrary, if $\frac{\partial Q_{tot}}{\partial Q_a} < 0$, then we should decrease the performance $J(\pi^a)$ of agent $a$, which means that we should perform gradient-descent algorithm:

$$\theta_{t+1} = \theta_t - \alpha\gamma^t \sum_u \nabla \pi(u|\tau_t^a) Q^\pi(\tau_t^a, u) \qquad (12)$$

In a word, Equation 11 and 12 can be rewrote as

$$\theta_{t+1} = \theta_t + \alpha \, sgn\left(\frac{\partial Q_{tot}}{\partial Q_a}\right) \gamma^t \sum_u \nabla \pi(u|\tau_t^a) Q^\pi(\tau_t^a, u) \qquad (13)$$

Similarly, for the continuous action space setting, we have

$$\theta_{t+1} = \theta_t + \alpha \, sgn\left(\frac{\partial Q_{tot}}{\partial Q_a}\right) \gamma^t \nabla_\theta \mu_\theta(\tau_t^a) \nabla_u Q^\mu(\tau_t^a, u)|_{u=\mu_\theta(\tau_t^a)} \qquad (14)$$

In this way, we can remove the monotonicity constraint. Figure 1 illustrates the overall setup.

For each agent $a$, we use DRQNs to encode per-agent observation $o_t^a$ and last action $u_{t-1}^a$ into a hidden state $h_t^a$ representing the action-observation history $\tau_t^a$. And then the hidden state $h_t^a$ is not only passed into an MLP estimating per-agent action-value, but also passed into a parameterized policy outputting the probability of available actions, as show in Figure 1b. If we plan to reuse the DRQN for each agent, we need to encode the "agent id" into the input of DRQN, and then the input changes to $(o_t^a, u_{t-1}^a, a)$.

The mixing network of QMIX is replaced by a MLP estimating the joint action-value and the extra state $s_t$ is also directly passed into the MLP because there is no monotonicity constraint, as show in Figure 1a.

Note that agents can reuse the critic network but can't reuse the actor policy network. Because $\sum_u \nabla \pi^a(u|\tau_t^a, \boldsymbol{\theta}) Q_a(\tau_t^a, u)$ is local for the agent $a$. If the term is mixed and is reused by all agents, then the $\sum_u \nabla \pi^a(u|\tau_t^a, \boldsymbol{\theta}) Q_a(\tau_t^a, u)$ and $\sum_a \sum_u \nabla \pi^a(u|\tau_t^a, \boldsymbol{\theta}) Q_a(\tau_t^a, u)$ may have opposite signs, which will affect the gradient-

ascent or gradient-descent algorithm.

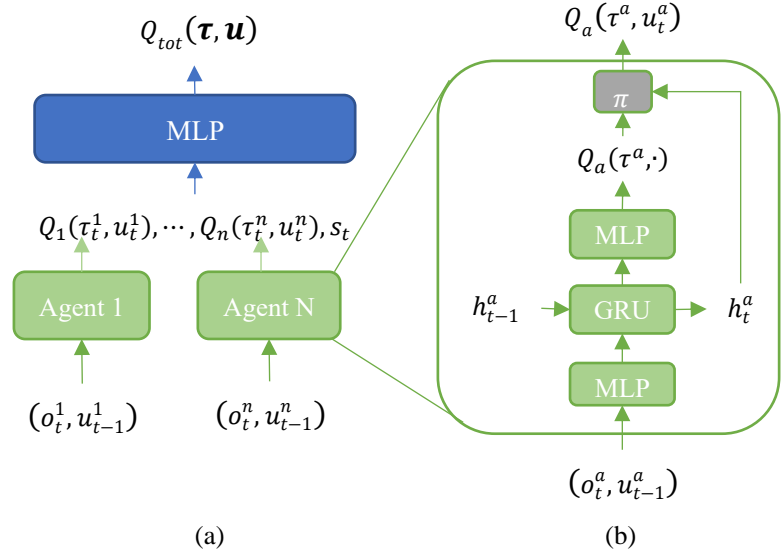

(a) (b)

*Figure* 1: (a) The overall architecture of NQMIX. (b) Agent network structure

### 4.3 Algorithm

**Algorithm 1** NQMIX
---
Input: an evaluate policy parameterization $\pi^a(u|\tau, \boldsymbol{\theta})$ for each agent $a$
Input: a target policy parameterization $\pi^a(u|\tau, \boldsymbol{\theta}')$ for each agent $a$
Input: an evaluate agent network reused by all agents
Input: a target agent network reused by all agents
Input: an evaluate mixing network
Input: a target mixing network
1: Loop
2:    Generate an episode following: $u_t^a \sim \pi^a(\cdot | \tau_t^a, \boldsymbol{\theta})$ $\forall a$
3:    Add the episode to replay buffer
4:    Sample a random mini-batch of $N$ episodes from replay buffer
5:    $I \leftarrow 1$
6:    **for** each time step $t$:
7:        $Q_a(\tau_t^a, \cdot) \leftarrow$ Agent_eval$(o_t^a, u_{t-1}^a)$ $\forall a$
8:        $Q'_a(\tau^a, \cdot) \leftarrow$ Agent_target$(o_{t+1}^a, u_t^a)$ $\forall a$
9:        $V'_a(\tau_{t+1}^a) \leftarrow \sum_u \pi^a(u|\tau_{t+1}^a, \boldsymbol{\theta}') Q'_a(\tau_{t+1}^a, u)$ $\forall a$
10:       $Q_{tot} \leftarrow$ Mixing_eval$[Q_1(\tau_t^1, u_t^1), \cdots, Q_n(\tau_t^n, u_t^n), s_t]$
11:       $V'_{tot} \leftarrow$ Mixing_target$[V'_1(\tau_{t+1}^1), \cdots, V'_n(\tau_{t+1}^n), s_{t+1}]$
12:       $\delta \leftarrow R + \gamma V'_{tot} - Q_{tot}$
13:       Update critic by minimizing the TD-error: $\delta$
14:       Update the actor policy:
$$\boldsymbol{\theta} \leftarrow \boldsymbol{\theta} + \alpha I \cdot \text{sign}\left(\frac{\partial Q_{tot}}{\partial Q_a}\right) \sum_u \nabla \pi^a(u|\tau_t^a, \boldsymbol{\theta}) Q_a(\tau_t^a, u) \quad \forall a$$

15:       $I \leftarrow \gamma I$
16:       Update the target networks:

$$\boldsymbol{\theta}' \leftarrow \tau\boldsymbol{\theta} + (1-\tau)\boldsymbol{\theta}'$$
$$\boldsymbol{w}'_{agent} \leftarrow \tau\boldsymbol{w}_{agent} + (1-\tau)\boldsymbol{w}'_{agent}$$
$$\boldsymbol{w}'_{mixer} \leftarrow \tau\boldsymbol{w}_{mixer} + (1-\tau)\boldsymbol{w}'_{mixer}$$

17:     **end for**
18: end Loop

A new sampled episode will be put int replay buffer, and NQMIX sample a random min-batch from replay buffer. NQMIX learned off-policy, so it can use previous experience to train current actor and critic and mini-batch gradient descent and ascent make each update more stable. The ability to use the replay buffer is one of the reasons why NQMIX performs better than COMA.

The learning target of NQMIX is state value (9$^{th}$ code), which can avoid overestimating learning target.

NQMIX uses soft update rather than replace target networks directly when updating target networks (16$^{th}$ code).

---

**Algorithm 2** NQMIX for continuous action space

---

Input: an evaluate policy parameterization $\mu^a(\tau|\boldsymbol{\theta})$ for each agent $a$
Input: a target policy parameterization $\mu^a(\tau|\boldsymbol{\theta}')$ for each agent $a$
Input: an evaluate agent network reused by all agents
Input: a target agent network reused by all agents
Input: an evaluate mixing network
Input: a target mixing network
1: Loop
2:     Generate an episode following: $u_t^a = \mu^a(\tau|\boldsymbol{\theta})\ \forall a$
3:     Add the episode to replay buffer
4:     Sample a random mini-batch of $N$ episodes from replay buffer
5:     $I \leftarrow 1$
6:     **for** each time step $t$:
7:       $Q_a(\tau_t^a,\cdot) \leftarrow \text{Agent\_eval}(o_t^a, u_{t-1}^a)\ \forall a$
8:       $Q'_a(\tau^a,\cdot) \leftarrow \text{Agent\_target}(o_{t+1}^a, u_t^a)\ \forall a$
9:       $Q_{tot} \leftarrow \text{Mixing\_eval}[Q_1(\tau_t^1, u_t^1), \cdots, Q_n(\tau_t^n, u_t^n), s_t]$
10:     $Q'_{tot} \leftarrow \text{Mixing\_target}[Q'_1(\tau_{t+1}^1, \mu^1(\tau_{t+1}^1|\boldsymbol{\theta}')), \cdots, Q'_n(\tau_{t+1}^n, \mu^n(\tau_{t+1}^n|\boldsymbol{\theta}')), s_{t+1}]$
11:     $\delta \leftarrow R + \gamma Q'_{tot} - Q_{tot}$
12:     Update critic by minimizing the TD-error: $\delta$
13:     Update the actor policy:

$$\boldsymbol{\theta} \leftarrow \boldsymbol{\theta} + \alpha I \cdot \text{sgn}\left(\frac{\partial Q_{tot}}{\partial Q_a}\right) \nabla_\theta \mu^a(\tau_t^a|\boldsymbol{\theta}) \nabla_u Q_a(\tau_t^a, u)|_{u=\mu^a(\tau_t^a)}\ \forall a$$

14:     $I \leftarrow \gamma I$
15:     Update the target networks:

$$\boldsymbol{\theta}' \leftarrow \tau\boldsymbol{\theta} + (1-\tau)\boldsymbol{\theta}'$$
$$\boldsymbol{w}'_{agent} \leftarrow \tau\boldsymbol{w}_{agent} + (1-\tau)\boldsymbol{w}'_{agent}$$

$$w'_{mixer} \leftarrow \tau w_{mixer} + (1-\tau) w'_{mixer}$$

16:   **end for**
17: end Loop

In addition, QMIX is difficult to learn in a continuous action space setting, but NQMIX can adapt to such settings, as show in Algorithm 2. At this time, NQMIX adopts deterministic policy gradient and the policy is denoted as $\mu^a(\tau|\boldsymbol{\theta}): T \rightarrow U$. And the learning target and the actor's updating formula need to be modified accordingly.

## 5  Experimental Results

### 5.1  Main Results

We test our algorithm on StarCraft II maps from the SMAC benchmark[18]. We pause training every 5,000 time-steps and run 32 evaluation episodes with learned parameterized policy or greedy policy. After training, we plot the mean win rate across 3 runs for each method on selected maps. Appendix B contains additional experimental details. We compare our NQMIX, QMIX and COMA on several SMAC maps. And in all maps, we set the difficulty to 7(very hard).

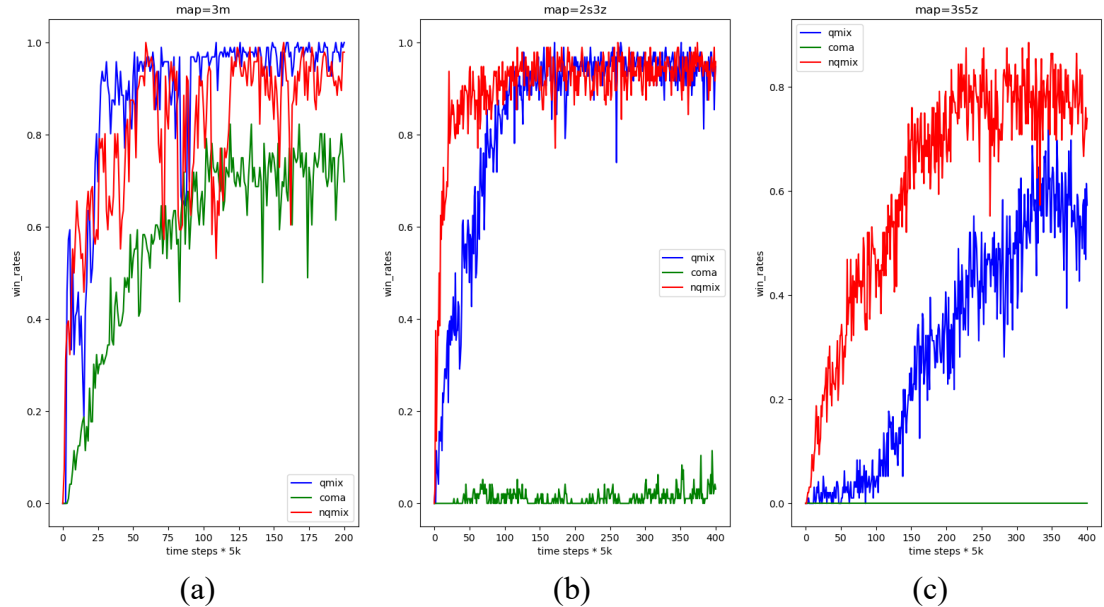

*Figure* 2. Win rates for QMIX, COMA and NQMIX on three different combat maps.

Figure 2(b) and 2(c) show the win rates of NQMIX on complex maps are better than QMIX and COMA while Figure 2(a) show the win rates of NQMIX on simple maps are between QMIX and COMA. Although NQMIX and QMIX have different mixers, yet the number of parameters is similar for both mixers. As the number of parameters in mixer gradually increases from Figure 2(a) to Figure 2(b), NQMIX will outperform QMIX, especially on the maps with complex heterogeneous agent types.

## 5.2 Ablation Results

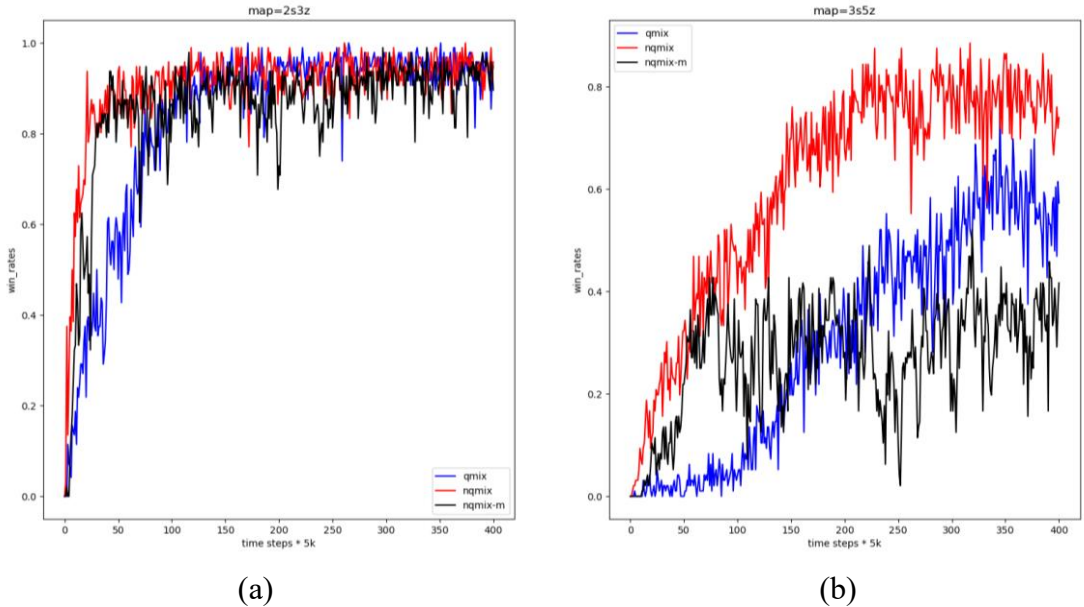

(a)                                                                           (b)

Figure 3. Win rates for QMIX, NQMIX, NQMIX-M on two different combat maps.

The ablation experiment is to explore whether the NQMIX should directly replace the original mixer by MLP. First, a method called NQMIX-M is added in this section and NQMIX-M continues to use the mixer of QMIX but remove the absolute activation function of mixer.

Figure 3a shows that NQMIX-M has slightly lower performance than NQMIX but still learns faster than QMIX. Figure 3b shows that NQMIX-M performs worse than both NQMIX and QMIX. Ablation results show that NQMIX can obtain better performance by modifying the mixer.

## 6 Conclusion and Future work

In this paper, we propose a novel method called NQMIX that can implement non-monotonic value function factorization by introducing an actor on QMIX. Our results in SMAC domain show that NQMIX has a stronger performance than QMIX and COMA on maps with complex heterogeneous agent types. In the future work, we aim to evaluate NQMIX on additional maps from SMAC to observe whether NQMIX will be have a satisfactory performance on some other maps and whether the actor introduced will become a new performance bottleneck.

# 7  Acknowledgements

## A. Solutions

My solutions to exercise 13.2 of [9]:
From Equation 3, we have

$$\nabla_\theta J(\pi) = \sum_s d^b(s) \sum_u \nabla_\theta \pi(u|s, \boldsymbol{\theta}) Q^\pi(s, u)$$

$$= \sum_s \sum_{t=0}^\infty \gamma^t P\{S_t = s|s_0, b\} \sum_u \nabla \pi(u|s) Q^\pi(s, u)$$

$$= \sum_{t=0}^\infty \gamma^t \sum_s P\{S_t = s|s_0, b\} \left[ \sum_u \nabla \pi(u|s) Q^\pi(s, u) \right]$$

$$= \sum_{t=0}^\infty \gamma^t E_s \left[ \sum_u \nabla \pi(u|s) Q^\pi(s, u) \right]$$

## B. Experimental Setup

### B.1. Architecture of NQMIX

All agents reuse a Deep Recurrent Q-Networks which consist of a 64-dimensional fully-connected layer, a GRU recurrent layer and a $|U|$-dimensional fully-connected layer. In other words, this part of NQMIX is the same as QMIX. But each agent has its local policy that is a 64-dimensional fully-connected layer with ReLU activation before it. Each local policy takes the hidden state of GRU as input. In order to make the numbers of parameters of mixers for both NQMIX and QMIX similar, the mixer of NQMIX is a MLP consisting of two fully-connected layers where the number of units of first layer is $32 \times (A + 4) - A$ ($A$ is the number of agents) and the second layer has one unit.

We used $\tau = 0.001$ for the soft target updates and $\gamma = 0.99$ for a discount factor. We use RMSprop for learning critic parameters with a learning rate of $5 \times 10^{-4}$. We use RMSprop for learning each actor policy parameters with a learning rate of $5 \times 10^{-4}$.